\documentclass[10pt,twocolumn,letterpaper]{article}

\usepackage{cvpr}
\usepackage{times}
\usepackage{epsfig}
\usepackage{graphicx}
\usepackage{amsmath}
\usepackage{amssymb}

\usepackage[breaklinks=true,bookmarks=false]{hyperref}

\cvprfinalcopy 

\def\cvprPaperID{****} 

\begin{document}

\title{Action sequence learning for causal shape transformation}

\author{Kin Gwn Lore* \and Daniel Stoecklein* \and Michael Davies$^\dag$ \and Baskar Ganapathysubramanian* \and Soumik Sarkar* \\
*Department of Mechanical Engineering, $^\dag$ Department of Electrical and Computer Engineering\\
Iowa State Universty\\
{\tt\small (kglore, stoeckd, mdavies, baskarg, soumiks)@iastate.edu}
}

\maketitle

\begin{abstract}

Deep learning became the method of choice in recent year for solving a wide variety of predictive analytics tasks. For sequence prediction, recurrent neural networks (RNN) are often the go-to architecture for exploiting sequential information where the output is dependent on previous computation. However, the dependencies of the computation lie in the latent domain which may not be suitable for certain applications involving the prediction of a step-wise transformation sequence that is dependent on the previous computation only in the visible domain. We propose that a hybrid architecture of convolution neural networks (CNN) and stacked autoencoders (SAE) is sufficient to learn a sequence of actions that nonlinearly transforms an input shape or distribution into a target shape or distribution with the same support. While such a framework can be useful in a variety of problems such as robotic path planning, sequential decision-making in games, and identifying material processing pathways to achieve desired microstructures, the application of the framework is exemplified by the control of fluid deformations in a microfluidic channel by deliberately placing a sequence of pillars. Learning of a multistep topological transform has significant implications for rapid advances in material science and biomedical applications.
\end{abstract}

\section{Introduction}
Hierarchical feature extraction using deep neural networks has been very successful in accomplishing various tasks such as object recognition~\cite{HY08}, speech recognition~\cite{HDY12}, deep vision enhancement~\cite{lore2015llnet}, multi-modal sensor fusion~\cite{SS12}, prognostics~\cite{sarkar2015early,akintayo2016early}, engineering design~\cite{lore2015hierarchical}, policy reward learning~\cite{lore2015deep}, and relating DNA variants to diseases~\cite{LXLF14}. This paper proposes an application of deep learning by fusing multiple architectures in solving design engineering problems, which potentially accelerates the development of various fields such as manufacturing, chemical engineering, and biology.

\begin{figure*}[htb!]
\centering
\includegraphics[width=0.7\textwidth,trim={20 10 10 10}]{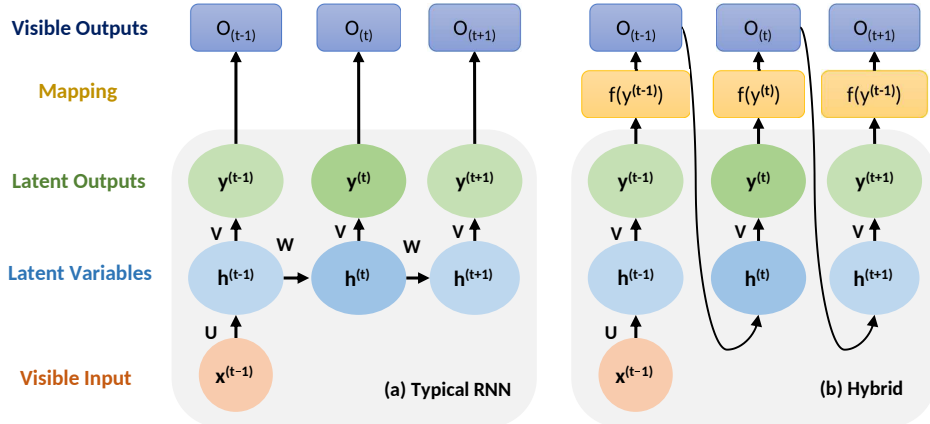} \caption{Difference between (a) RNN and the (b) hybrid approach. Both models take in input vector $\mathbf{x}$ at time $t$, pass it through the hidden layer $\mathbf{h}$ and produces a latent output $\mathbf{y}$. The latent outputs $\mathbf{y}$ are mapped to visible outputs $\mathbf{o}$ via function $\mathbf{f(\cdot)}$. In RNN, there are dependencies in the latent layer. In our hybrid approach for the considered problem, the visible output is used as the input to the next (i.e. dependent on the visible outputs) without dependencies in the latent layer.}
\label{fig:rnnvsaction}
\end{figure*}

For sequence prediction, recurrent neural networks (RNN)~\cite{mikolov2010recurrent} are often the go-to architecture for exploiting sequential information where the output is dependent on previous computation. However, the dependencies of the computation lie in the latent domain which may not be suitable for certain applications involving the prediction of a step-wise transformation sequence that is dependent on the previous computation only in the visible domain~(Fig.~\ref{fig:rnnvsaction}). In the paper, we propose a deep learning architecture which simultaneously predicts the intermediate shape between two images and learns a sequence of causal actions contributing to desired shape transformation in the visible domain. This topological transformation framework has multiple implications. The architecture can be easily implemented in applications such as learning to transform the belief space for robotic path planning~\cite{zhang2015learning}, sequential decision making in games~\cite{silver2016mastering}, learning the material processing pathways to obtain desired microstructures starting from an initial microstructure~\cite{wodo2015automated}, and learning a sequence of manufacturing steps in additive manufacturing~\cite{Paulsen2015}, with \textit{fast-design} being the main advantage. The contributions of the paper are outlined next:
\begin{enumerate}
 \item A formulation of learning causal shape transformations to predict a sequence of transformation actions is presented, in the setting where only an initial shape and a desired target shape are provided.
 \item An integrated hierachical feature extraction approach using stacked autoencoders (SAE)~\cite{vincent2008extracting} with convolutional neural networks (CNN)~\cite{KSG12} is proposed to capture transformation features to generate the associated sequence resulting in the transformation.
 \item The proposed approach is tested and validated via numerical simulations on an engineering design problem (i.e. flow sculpting in microfluidic devices), with results showing competitive prediction accuracy over previously explored methods.
\end{enumerate}

\section{Problem setup and previous approach}\label{sec:problem}
In this section, we describe the problem setup and provide a brief background on the previous approach. To illustrate how the previous and current architectures can be implemented, we will focus our attention on \textit{microfluidic flow sculpting} as the application.

\subsection{Microfluidic flow sculpting}

\begin{figure*}[htb!]
\centering
\includegraphics[width=\textwidth,trim={0 0 0 0}]{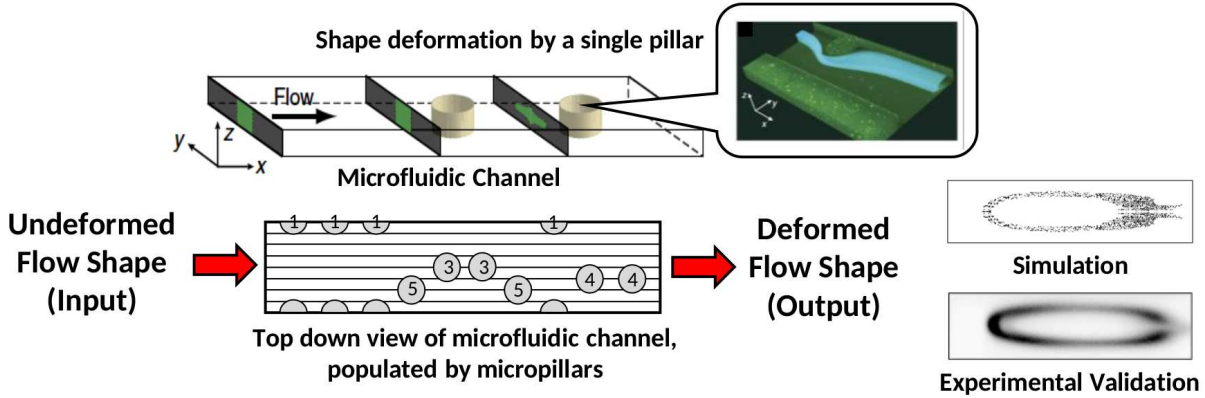} \caption{Pillar programming. Each pillar contributes to the deformation of the flow. The position-diameter pair of each pillar is assigned an index which will be used as class labels for classification.}
\label{fig:fig_problemformulation}
\end{figure*}

Inertial fluid flow sculpting via micropillar sequences is a recently developed method of fluid flow control with a wealth of applications in the microfluidics community~\cite{ASM13}. This technique utilizes individual deformations imposed on fluid flowing past a single obstacle (pillar) in confined flow to create an overall net deformation from a rationally designed sequence of such pillars. If the pillars are spaced far enough apart (space $>6D$, for a pillar diameter $D$), the individual deformations become independent, and can thus be pre-computed as building blocks for a highly efficient forward model for the prediction of sculpted flow given an input pillar sequence~\cite{SWD15} (Fig.~\ref{fig:fig_problemformulation}). Since its debut, flow sculpting via micropillar sequence design has been used in fundamental applications of novel particle fabrication~\cite{Paulsen2015}, and solution transfer~\cite{Sollier2015}.

However, practical applications require solving the \textit{inverse problem}, that is, to generate a sequence of pillars given a user-defined flow shape. Without intelligent computer algorithms, such tasks require time-consuming trial and error design iterations. The automated determination of pillar sequences that yields a custom shape is an impactful advance. Although researchers have tried to frame this inverse problem as an unconstrained optimization problem~\cite{SWD15}, they are invariable time-consuming. While many methods are used to solve the forward problem~\cite{MMK99,DPCNDBSA14,BKE10,AFJH95}, only a limited amount of effort has been done in solving the inverse problem~\cite{lore2015hierarchical}. For practical and time-efficient applications, deep learning methods are explored to map user-defined flow shapes and the corresponding pillars of sequence. Additionally, this application serves as a very impactful advance towards learning topological transformations.

This framework can be utilized for concrete biomedical applications that require the design of microfluidic devices. Possible applications include: (a) designing a device to move fluid surrounding cells (e.g. lymphoid leukemia cells) against a far wall of the microfluidic channel where it can be collected at high purity while the cells are maintained at the channel center. High purity allows the reuse of valuable reagents for staining cells during diagnosis, (b) wrapping a fluid around the microchannel surface to characterize binding p24 (an HIV viral capsid protein) to anti-p24 antibody immobilized on the microchannel surface. Flow sculpting can enhance reaction of low abundance proteins that can improve diagnostic limits of detection for various diseases. Hence, this study may promise new application areas for the machine learning community related to thermo-fluid sciences and design engineering.

\subsection{Discretization of the design space}
To approach the inverse problem, class indices are assigned to pillars with different specifications. For instance, a pillar located at position $0.0$ with a diameter of $0.375$ is an index of 1, whereas another pillar at position $0.125$ and diameter $0.375$ is assigned an index of $2$. Diameter and position values are represented as ratios with respect to the channel size and locations, so they are dimensionless quantities to help enable scalability of the fluid channel. Index assignment is performed over a finite combination of pillar positions and diameters that has been obtained by discretizing the design space. In the study, there are 32 possible indices (or classes) that describe the diameter and position of a single pillar.

\subsection{Previous approach: Simultaneous multi-class classification (SMC)}

Simultaneous multi-class classification is a method used in~\cite{lore2015hierarchical} to solve a similar problem. Instead of solving a single classification problem, the model solves a sub-problem for each pillar using the parameters learned by the CNN. This formulation requires a slight modification in the loss function: for pillar sequence with length $n_p$, the loss function to be minimized for a data set $\mathcal{D}$ is the negative log-likelihood defined as:
\[ \ell_{total}(\theta,\mathcal{D}) = - \sum_{j=1}^{n_p} \sum_{i=0}^{\mathcal{|D|}} \left[ \log{\left(P(Y=y^{(i)}|x^{(i)},W,b)\right)} \right]_j \]
where $\mathcal{D}$ denotes the training set, $\theta$ is the model parameters with $W$ as the weights and $b$ for the biases, $y$ is predicted pillar index whereas $x$ is the provided flow shape image. The total loss is computed by summing the individual losses for each pillar.

\begin{figure*}[!htb]\vspace{-5pt}
\centering
\includegraphics[width=\textwidth,trim={0 0 0 0}]{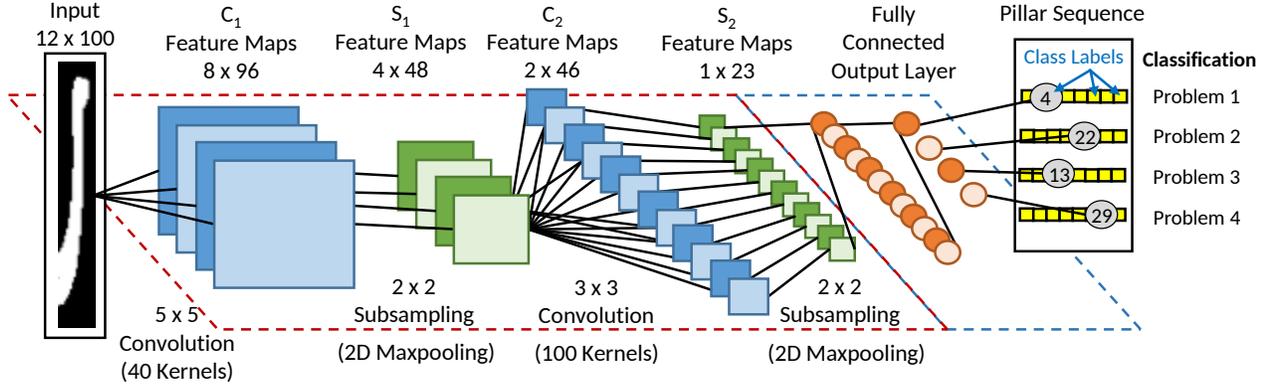} \caption{CNN with the SMC problem formulation (previous approach). A classification problem is solved for each pillar in the sequence.}
\label{fig:fig_cnnmc}\vspace{-10pt}
\end{figure*}

\section{Action sequence learning for causal shape transformation}\label{sec:aseq}
The CNN-SMC framework in~\cite{lore2015hierarchical} is not without drawbacks. The pillars are predicted in a joint manner, where each of the pillars in the sequence are inferred simultaneously. While it may produce satisfactory sequences which regenerates the desired shape, it is unclear how the successive pillars interact with one another. In this section, we propose a deep learning architecture which predicts the intermediate shape between two images and learns a sequence of causal actions contributing to desired shape transformation. The architecture can be easily implemented in applications such as:

\begin{enumerate}
\item Learning to transform the belief space for robotic path planning. A robot may be chasing a mobile target while maintaining a posterior (that is transformed into a visual representation analogous to the flow shape) corresponding to the location of the target. If we would like to specifically maximize posterior in a certain region, the corresponding problem would be: ``What is the best course of action that should be taken by the robot in succession?".
\item Learning the material processing pathways to obtain the desired microstructures. In materials processing, processing the materials by altering the properties in succession will alter the morphology (microstructures) of the material. The equivalent inverse problem would be: ``If we want the material to achieve a specific morphology, what are the processing steps that should be taken?".
\item Learning a sequence of manufacturing steps in additive manufacturing, with \textit{fast-design} being the main advantage.
\end{enumerate}

While CNN-SMC is capable of predicting a large number of different sequences in a total time of just seconds, a drawback is that the sequence length is constrained because a new model needs to be trained to generate a sequence with different lengths. Furthermore, the sequence which deforms into the target flow shape is predicted in a joint manner and do not provide sufficient insight on the interplay between pillars causing the deformation.

RNN-like architectures are, although scalable, deemed unsuitable because the elements in the output vector in our problem are generally independent of each other, unlike words in a sentence. In this application, a pillar can be \textit{independently} added to the flow channel only after the flow shape resulted from the previous pillar stabilizes, i.e. no longer changes. This way, we guarantee that there are no interactions among pillars (hence the independence). However, all these pillars contribute to the change in the \textit{overall} flow shape (thus they are causal to the final shape). In the context of RNN outputs, e.g. for the case of next-word prediction, oftentimes the output vectors are not independent-one word will have a high probability to be the next word of a certain word. In our context, the choice of the next pillar will not depend on the choice of the previous pillar. Rather, we look at the \textit{now} by making use of the given generated shape so far regardless of the history.

A related concept is the spatial transformer network~\cite{jaderberg2015spatial} where the localization network outputs geometrical transformation parameters; however we desire an exact class attributed to an arbitrary transformation function. In this work, we predict the pillar sequence one pillar at a time without disregarding causality, such that the produced sequences deform the flow into one that resembles more closely to the desired shape. 

\section{Proposed architectures}\label{sec:APNitn_proposed}

In this section, we present two core architectures and the integrated framework which fuses these two architectures.

\subsection{Action prediction network (APN)}

\begin{figure*}[!htb]
\centering
\includegraphics[width=\textwidth,trim={0 0 0 0}]{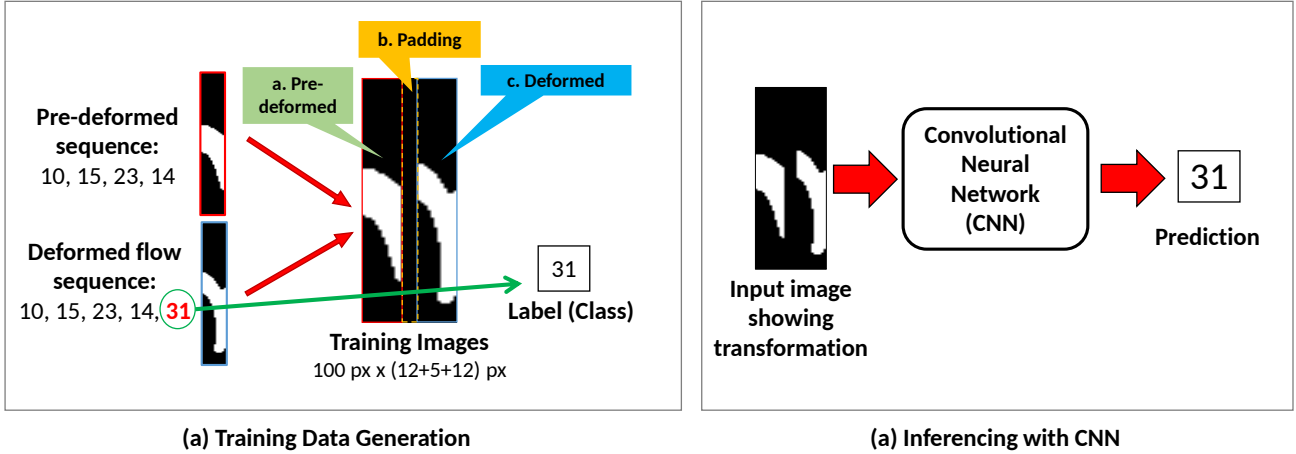} \caption{The APN addresses the question: `Given a pair of pre- and post-deformed shape, what is the identity of the pillar causing the deformation?'}
\label{fig:fig_APN}
\end{figure*}

To predict the sequence of pillars that results in the desired flow shape, this chapter introduces the notion of \textit{transformation learning}. Fig.~\ref{fig:fig_APN} shows the learning approach by supplying juxtaposed flow shapes, one before deformation, and one after, into a CNN that extracts relevant features and predicts the class of the pillar causing the deformation. In the training data generation procedure, the input is comprised of three parts: (a) the pre-deformed flow shape, which is produced from a randomly generated sequence of varying length with values up to the number of classes of a single pillar; (b) a padding to prevent the convolutional kernels to pick up interfering features from the juxtaposed images; and (c) the post-deformed shape produced from adding a random pillar index to the previous sequence. This newly-added pillar index will become the label to train the CNN. Essentially, the action prediction network predicts the index of the pillar (which describes its position and diameter in the flow channel) given a pre- and post-deformed shape.

When actually inferencing, the right portion (part c) of the input image is replaced by the final target shape. The input image is fed into the CNN to obtain a pillar index, which is added to an initially empty pillar sequence. Because the forward model (i.e. going from sequence to flow shape) is not at all computationally expensive and takes merely milliseconds, this pillar sequence is used to regenerate the \textit{current shape} which replaces the left portion (part a) of the input, while the right portion (part c, the target) remains the unchanged. The input enters the CNN again to obtain a second pillar index, and is subsequently added to the previously obtained pillar sequence. At this point, we have a sequence of length 2, where the sequence will then again used to regenerate a new current shape. The process is repeated until the \textit{current shape} matches the \textit{target shape}, or until a user-defined stopping criterion is met.

\begin{figure}[!htb]
\centering
\includegraphics[width=0.5\textwidth,trim={0 0 0 0}]{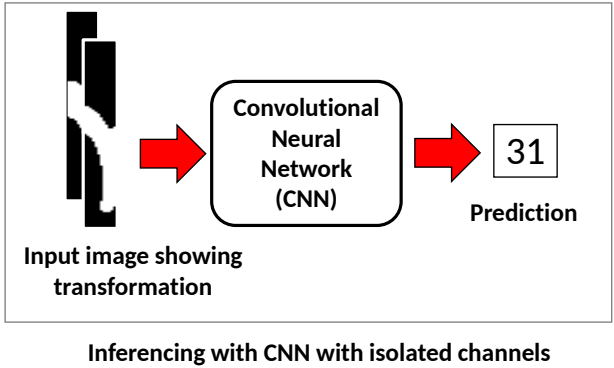} \caption{APN-C treats the pre- and post-deformed shapes as separate channels.}
\label{fig:fig_APNc}
\end{figure}

An alternative APN is to feed the pre- and post-deformed shapes into the CNN separately as \textit{isolated channels} before merging them together in the fully connected layer (Fig.~\ref{fig:fig_APNc}). This architecture will be referred to as APN-C (where the `C' denotes \textit{channels} and signifies that the pre- and post-deformed shapes are treated as different channels). In other words, each channel has different sets of filters and the relationship between the two sets will fuse together at the fully connected layer.

However, this method only works well for simpler target shapes. For more complex flow shapes (e.g. shapes with many sharp angles, jagged edges, swirls and curls; see Fig.~\ref{fig:fig_results3} sample 20A for example), the transformation path may be highly nonlinear-- the \textit{current shape} may never converge to the final desired shape. Furthermore, the training data covers a vanishingly small fraction of the design space with coverage shrinking exponentially as the sequence length increases (i.e., an $n_p$ sequence will result in $32^{n_p}$ different combinations), so it is necessary to learn the transformations in a meaningful way. To help alleviate this issue, we will introduce the Intermediate Transformation Network (ITN) in the next subsection.

\textbf{APN parameters: } The input image into the APN is comprised of two $12\times100$ px flow shape image with a $5\times100$ px padding in between, resulting in a final dimension of $29\times100$ px. For APN-C, the two inputs are treated as separate channels without introducing padding. The model contains two convolutional layers with 40 and 100 kernels respectively (sizes of $5\times 5$ and $3\times 3$ px), each followed by a $2\times2$ maxpooling layer. The fully connected layer has 500 hidden units. Training is done with 250,240 training examples and 60,000 validation examples in minibatches of 50 with a learning rate of 0.01. The training procedure employs the early stopping algorithm where training stops when validation error ceases to decrease.

\subsection{Intermediate transformation network (ITN)}

\begin{figure*}[!htb]
\centering
\includegraphics[width=\textwidth,trim={0 0 0 0}]{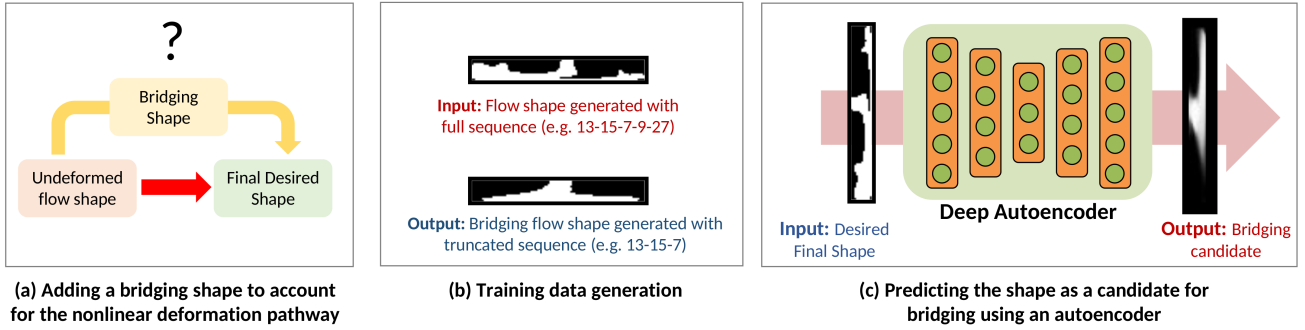} \caption{The ITN addresses the question: `Given the final shape, what is the possible shape that lies in the middle of the nonlinear deformation pathway?'}
\label{fig:fig_itn}
\end{figure*}

The ITN attempts to construct a flow shape that bridges between the purely undeformed flow shape (i.e. shape generated with an empty sequence) and the final desired flow shape in the nonlinear transformation path (See Fig.~\ref{fig:fig_itn}(a)). We used a deep autoencoder to extract hierarchical features from the desired final shape, and outputs an approximated \textit{bridging shape}. To generate the training data for this network, the input image is generated with a random pillar sequence with a varying length. The corresponding bridging shape is generated by truncating the same pillar sequence by half, thus the shape lies in the middle of the transformation pathway (Fig.~\ref{fig:fig_itn}(b)). Formally, if a sequence has length $n$, it is truncated to $(n+1)/2$ if $n$ is odd, and to $n/2$ if $n$ is even. This pair of images is used to train a deep autoencoder, where the mean squared error (MSE) between the model outputs and the desired bridging shape is backpropagated to finetune the weights and biases of the model.

During inference, the edges in the outputs of the ITN may appear blur because the bridging shape is only an approximation (Fig.~\ref{fig:fig_itn}(c)). To obtain binary images, we thresholded the pixel values.

The ITN may be extended to obtain several \textit{waypoints} instead of only the midpoint, where the number of waypoints is dictated by the complexity of the target flow shape (see section~\ref{sec:complex} for details). Doing so will allow a smoother transformation pathway, but will require some changes in the training data generation procedure. We call this the \textit{recursive ITN} (rITN) and leave this as a future work. Some may comment that ITN is somewhat similar to RNN. We note that for this problem (along with the example applications discussed in section~\ref{sec:aseq}) requires transforming the hidden states into the physical domain using a complex forward function before making the next prediction. Unlike RNNs, ITNs do not solely use the latent output directly as the input in the next time step. Due to the forward function, we need to constantly transform the output back into the physical domain and re-evaluate at every step. One may also argue that ITN is somewhat similar to RNN in terms of \textit{memory}, but by the same argument, the difference in domain is the key. If only the latent domain is used for every step, the final sequence output may not necessarily reflect the true transformation in the physical domain.

\textbf{ITN parameters: } The autoencoder has 3 layers of 500 hidden units each and accepts flow shape images of $12\times100$ px. Training was done on 500,000 training examples and 20,000 validation examples in minibatch of size 1,000 with a learning rate of 0.01. Training is also done using the early-stopping algorithm.

\subsection{The integrated pipeline (APN+ITN)}
\begin{figure*}[!htb]
\centering
\includegraphics[width=\textwidth,trim={0 0 0 0},clip]{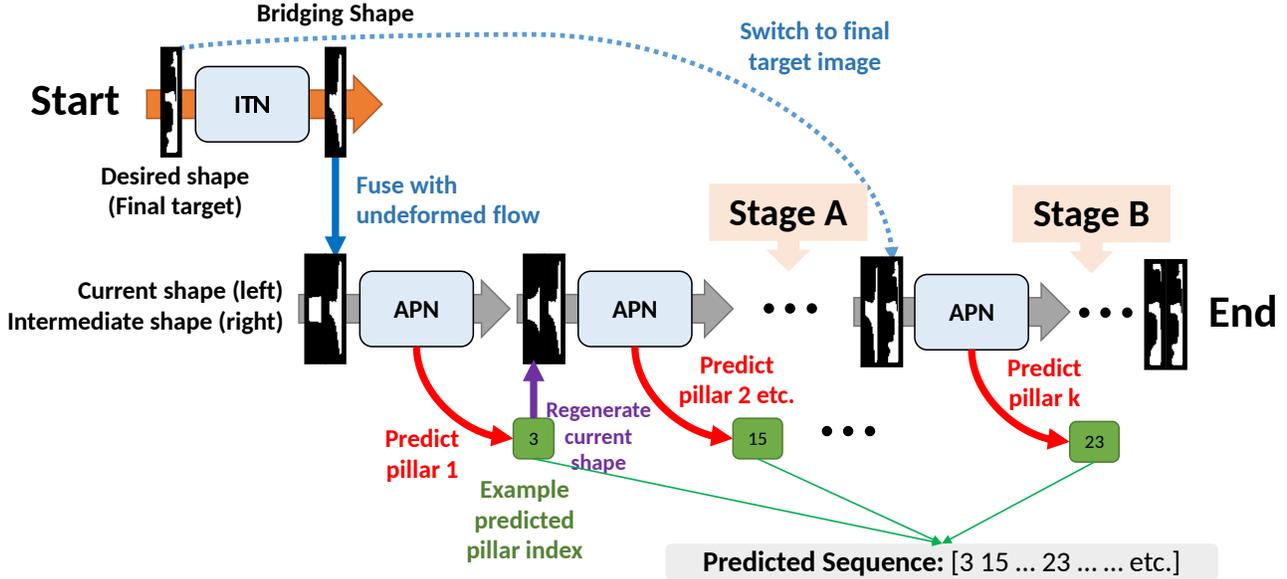} \caption{The integrated pipeline combining both APN and ITN.}
\label{fig:fig_integrated}
\end{figure*}

With the roles of the APN and ITN clearly described, we can now integrate both networks to form an integrated pipeline. A schematic is shown in Fig.~\ref{fig:fig_integrated}. At the very beginning, the ITN automatically guesses a candidate for the bridging flow shape, which is then considered as the \textit{temporary} target shape, and is concatenated with the undeformed shape placed to its left with a padding. The concatenated input is supplied into the APN to predict the first pillar causing the deformation, and then added to the sequence which was initially empty, hence resulting in a sequence of length 1. The current shape is regenerated with the updated 1-pillar sequence and replaces the left portion of the input image for the next APN iteration to obtain the second pillar index. The process is repeated as `Stage A' (with the bridging shape as a temporary target) until the current shape is sufficiently similar to the bridging shape, or until an iteration limit is reached. Then, the right portion of the input image (which was the bridging shape) is replaced with the final desired shape as the target. This process is continued as `Stage B' which undergoes the same process as `Stage A', except the target shape is now the final desired shape. After each iteration, the predicted pillar index is added to the sequence until the reconstructed flow shape matches the desired shape or a stopping criterion is achieved. The resulting sequence will vary in length for different desired shapes.

\section{Results and discussions}\label{sec:APNitn_results}
In this section we first present the evaluation metrics used in the study, then show the results comparing CNN-SMC against our method using APN and the APN+ITN hybrid architecture.

\subsection{Complexity measure and image quality metric}\label{sec:complex}

\begin{figure*}[!htb]
\centering
\includegraphics[width=0.7\textwidth,trim={0 0 0 0},clip]{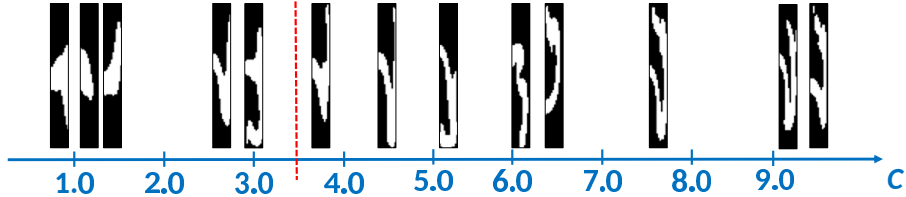} \caption{Perimetric complexity $C$ for different flow shape examples. Target flow shapes with $C>3.5$ are considered for testing.}
\label{fig:fig_complex}
\end{figure*}

The perimetric complexity, $C$ of binary digital images can be described by the following expression proposed by Attneave et al.~\cite{attneave1956quantitative}:
\[C = \frac{P^2}{4\pi A}\]
where $P$ is the perimeter of the shape, and $A$ is the area of the shape. This measure can be utilized to determine the number of intermediate shapes that should be predicted by the ITN. The higher the shape complexity, the more necessary are the use of bridging shapes. Fig.~\ref{fig:fig_complex} shows the complexity for various flow shapes.

To quantify the effectiveness of CNN-SMC versus our proposed approach, we evaluated the \textit{pixel match rate}, PMR on 20 target flow shapes and computed appropriate statistics. The PMR, defined in~\cite{lore2015hierarchical}, is computed as follows:
\[ \mbox{PMR} = 1 - \frac{|| \boldsymbol{p} - \boldsymbol{\hat{p}} ||_1}{|\boldsymbol{p}|} \]
where $\boldsymbol{p}$ is the target image vector, $\boldsymbol{\hat{p}}$ is the predicted image vector, and $|\boldsymbol{p}|$ denotes the cardinality of the vector (i.e., the total number of pixels in the image). Since shape transformation is an essential aspect in this problem formulation, the \textit{structural similarity index} (SSIM)~\cite{wang2004image} is used as a supplementary metric to compare how structurally similar are the regenerated flow shape images (from predicted sequence) to the target flow shape. SSIM explores the change in image structure and incorporates pixel inter-dependencies. SSIM is expressed as:
\[\hbox{SSIM}(x,y) = \frac{(2\mu_x\mu_y + c_1)(2\sigma_{xy} + c_2)}{(\mu_x^2 + \mu_y^2 + c_1)(\sigma_x^2 + \sigma_y^2 + c_2)}\]
where $\mu_x$ is the average of window $x$, $\mu_y$ is the average of window $y$, $\sigma_x^2$ is the variance of $x$, $\sigma_y^2$ is the variance of $y$, $\sigma_{xy}^2$ is the covariance of $x$ and $y$, $c_1=(k_1 L)^2$ and $c_2=(k_2 L)^2$ are two variables to stabilize the division with weak denominator with $k_1=0.01$ and $k_2=0.03$ by default, and $L$ is the dynamic range of pixel values.

\subsection{Predicting sequences with APN and APN+ITN}

\begin{figure*}[!htb]
\centering
\includegraphics[width=\textwidth,trim={0 0 0 0}]{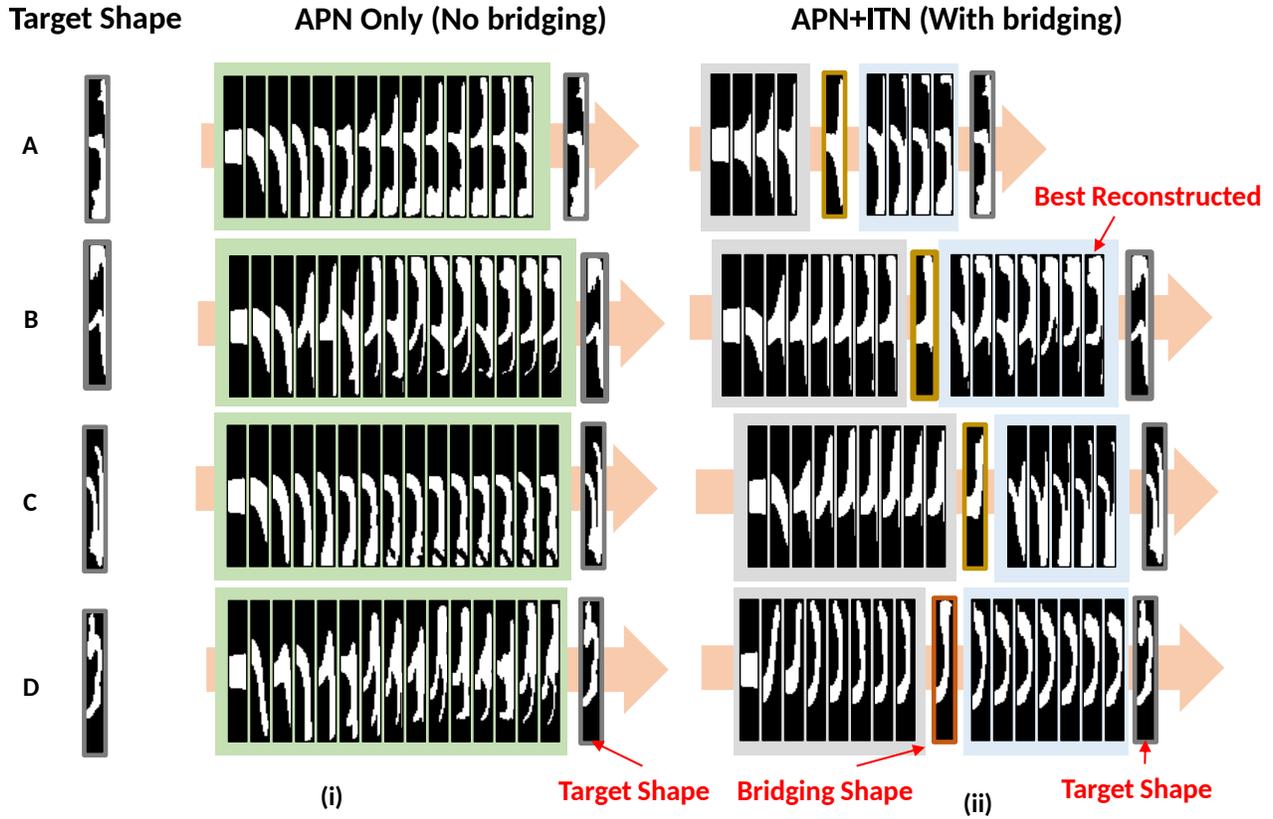} \caption{Four examples of sequence prediction using (i) APN-only without bridging and (ii) APN+ITN with bridging. By predicting a bridging shape, the resulting predicted sequence is able to reconstruct flow shapes that are more similar to the target shape. Each frame shows the deformation on the flow shape with each additional predicted pillar added into the sequence.}
\label{fig:fig_results}
\end{figure*}

In all of our tests, the target flow shape is generated from a 10-pillar sequence which is sufficiently complex. Fig.~\ref{fig:fig_results} shows four example target shapes with the performance using only APN (without bridging) and APN+ITN (with bridging). In most cases, the APN-only formulation produces pillar sequences resulting in flow shapes that do not resemble the target shape as close as using APN+ITN. This can be clearly seen for cases C and D in Fig.~\ref{fig:fig_results}. On the other hand, by using the bridging shape as a temporary target, the prediction performance saw great improvements. In addition, we see that most shapes are able to converge to both the bridging shape as well as the target in the APN+ITN formulation. In realistic applications, the sequence may be stored in memory and post-processed to remove the redundant pillars, thus producing a shorter sequence which may increase financial savings during the manufacturing process.

\subsection{Comparison of CNN+SMC, APN, APN-C, and APN+ITN}

\begin{figure*}[!htb]
\centering
\includegraphics[width=\textwidth]{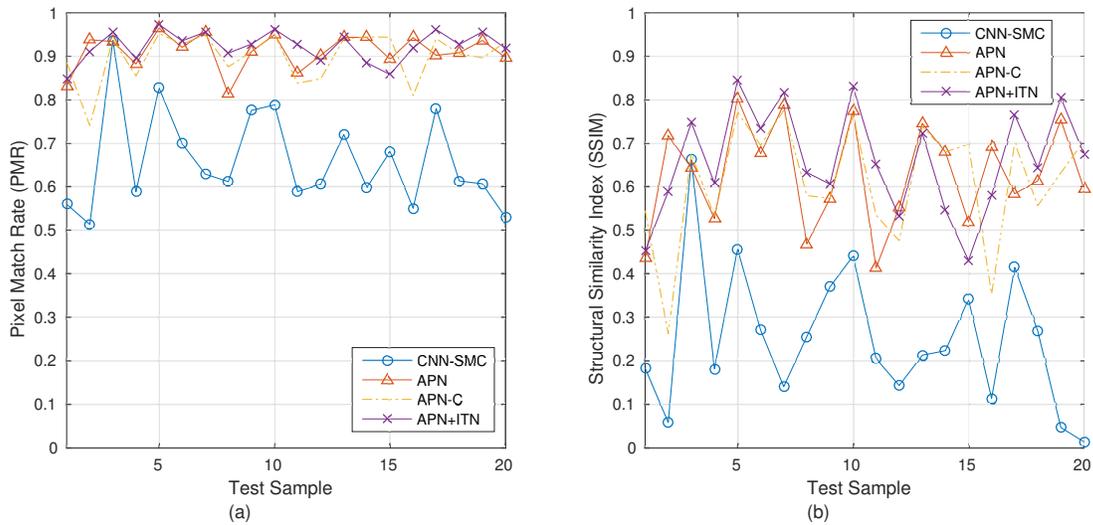} \caption{Sample-wise (a) PMR and (b) SSIM comparison using CNN+SMC, APN, APN-C, and APN+ITN. 20 test target shapes are randomly generated with a 10-pillar sequence.}
\label{fig:fig_results2}
\end{figure*}

\begin{figure*}[!htb]
\centering
\includegraphics[width=\textwidth,trim={0 0 0 0}]{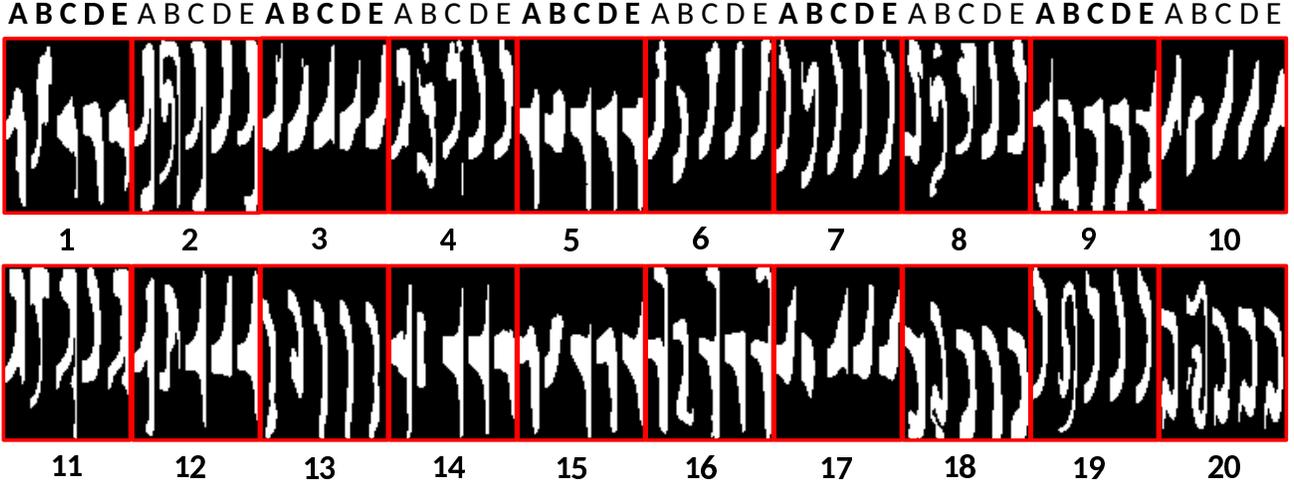} \caption{20 test shapes with reconstructed flow shapes generated from sequences predicted using different methods. Column A is the target shape, B for the reconstruction for CNN+SMC, C for APN, D for APN-C, and E for APN+ITN.}
\label{fig:fig_results3}
\end{figure*}

\begin{table*}[htb!]
\caption{PMR and SSIM of regenerated flow shapes using different enhancement methods. The numbers reported are in the format of [PMR/SSIM]. Bolded numbers correspond to the method with the highest PMR and SSIM. Asterisk (*) denotes our architecture presented in this paper.}
\begin{center}
{\small
\begin{tabular}{|l|c|c|c|c|c|}\hline
\textbf{Target Flow Shapes}    &	\textbf{CNN+SMC}	&	\textbf{APN*}	& \textbf{APN-C*} &	\textbf{APN+ITN*}	\\\hline
\textbf{Test Shape 1}	        &	0.5600	/	0.1829	&	0.8317	/	0.4349	&	\textbf{0.8867}	/	\textbf{0.5450} &	0.8475	/	0.4521	\\
\textbf{Test Shape 2}	        &	0.5133	/	0.0578	&	\textbf{0.9375}	/	\textbf{0.7160}	&	0.7408	/	0.2620	&	0.9117	/	0.5899	\\
\textbf{Test Shape 3}	        &	0.9367	/	0.6622	&	0.9342	/	0.6437	&	0.9400	/	0.6766	&	\textbf{0.9558}	/	\textbf{0.7481}	\\
\textbf{Test Shape 4}	        &	0.5883	/	0.1813	&	0.8833	/	0.5303	&	0.8550	/	0.5315	&	\textbf{0.8950}	/	\textbf{0.6082}	\\
\textbf{Test Shape 5}	        &	0.8275	/	0.4539	&	0.9650	/	0.8108	&	0.9525	/	0.7710	&	\textbf{0.9725}	/	\textbf{0.8407}	\\
\textbf{Test Shape 6}	        &	0.6992	/	0.2705	&	0.9208	/	0.6765	&	0.9283	/	0.6959	&	\textbf{0.9358}	/	\textbf{0.7346}	\\
\textbf{Test Shape 7}	        &	0.6300	/	0.1412	&	\textbf{0.9567}	/	0.7895	&	0.9500	/	0.7760	&	0.9558	/	\textbf{0.8167}	\\
\textbf{Test Shape 8}	        &	0.6117	/	0.2555	&	0.8125	/	0.4687	&	0.8758	/	0.5802	&	\textbf{0.9075}	/	\textbf{0.6330}	\\
\textbf{Test Shape 9}	        &	0.7767	/	0.3743	&	0.9117	/	0.5776	&	0.9108	/	0.5731	&	\textbf{0.9292}	/	\textbf{0.6105}	\\
\textbf{Test Shape 10}	        &	0.7883	/	0.4407	&	0.9500	/	0.7745	&	0.9508	/	0.7646	&	\textbf{0.9625}	/	\textbf{0.8299}	\\
\textbf{Test Shape 11}	        &	0.5892	/	0.2046	&	0.8608	/	0.4137	&	0.8383	/	0.5346	&	\textbf{0.9275}	/	\textbf{0.6511}	\\
\textbf{Test Shape 12}	        &	0.6058	/	0.1430	&	0.9025	/	\textbf{0.5527}	&	0.8483	/	0.4760	&	\textbf{0.8908}	/	0.5314	\\
\textbf{Test Shape 13}	        &	0.7200	/	0.2126	&	\textbf{0.9442}	/	\textbf{0.7458}	&	0.9400	/	0.7391	&	0.9433	/	0.7233	\\
\textbf{Test Shape 14}	        &	0.5983	/	0.2229	&	0.9433	/	0.6804	&	\textbf{0.9450}	/	\textbf{0.6804}	&	0.8850	/	0.5466	\\
\textbf{Test Shape 15}	        &	0.6808	/	0.3412	&	0.8933	/	0.5195	&	\textbf{0.9442}	/	\textbf{0.6980}	&	0.8583	/	0.4332	\\
\textbf{Test Shape 16}	        &	0.5508	/	0.1135	&	\textbf{0.9433}	/	\textbf{0.6919}	&	0.8100	/	0.3533	&	0.9200	/	0.5813	\\
\textbf{Test Shape 17}	        &	0.7792	/	0.4134	&	0.9033	/	0.5882	&	0.9408	/	0.7010	&	\textbf{0.9617}	/	\textbf{0.7655}	\\
\textbf{Test Shape 18}	        &	0.6133	/	0.2677	&	0.9075	/	0.6112	&	0.9058	/	0.5567	&	\textbf{0.9267}	/	\textbf{0.6416}	\\
\textbf{Test Shape 19}	        &	0.6075	/	0.0464	&	0.9367	/	0.7540	&	0.8967	/	0.6324	&	\textbf{0.9567}	/	\textbf{0.8044}	\\
\textbf{Test Shape 20}	        &	0.5300	/	0.0134	&	0.8983	/	0.6039	&	\textbf{0.9325}	/	\textbf{0.7051}	&	0.9183	/	0.6753	\\\hline
\textbf{Average}	            &	0.6603	/	0.2500	&	0.9118	/	0.6292	&	0.8996	/	0.6126	&	\textbf{0.9231}	/	\textbf{0.6609}	\\\hline
\end{tabular}
}
\end{center}
\label{tab:pmr_ssim_images}
\vspace{-15pt}\end{table*}

20 sample-wise comparison on the performance of CNN+SMC, and our methods APN and APN+ITN are shown in Fig.~\ref{fig:fig_results2} and tabulated in Table~\ref{tab:pmr_ssim_images}. The reconstructed flow shapes from the predicted sequences are shown in Fig.~\ref{fig:fig_results3}. In both Fig.~\ref{fig:fig_results2}(a) and Fig.~\ref{fig:fig_results2}(b), the PMR and SSIM for APN, APN-C, and APN+ITN are clearly higher than the CNN+SMC approach. We observe that for some target flow shapes, the predicted sequence for CNN+SMC may result in an entirely dissimilar shape (e.g. sample 2, 19, 20). In some cases (e.g. samples 14, 15, 16) APN fared better than the hybrid APN+ITN model. However, APN+ITN is consistently more superior in terms of both PMR and SSIM than the APN-only architecture (i.e. APN and APN-C). This shows that having a bridging shape generally helps in producing sequences that generate complex flow shapes. The inferior performance of CNN+SMC is due the model being constrained to always generate a sequence with 10 pillars, thus the resultant flow shapes often become overcomplicated.

Additionally, treating the input as separate channels results in lower performance while all other model parameters are kept equal. This suggests that having a single filter that learns features from both input images simultaneously is more effective in feature extraction compared to learning the features in isolation before merging at the fully connected layer.

A clear advantage of using both APN and ITN together is that the model does not need to be retrained for variable sequence lengths, unlike in the CNN-SMC model where the number of pillars in the output sequence is constrained. This method is highly scalable, and has an enormous room for extension into sculpting more highly complex flow shapes.

\section{Conclusions and Future Work}\label{sec:fs_conclusion}
This paper proposes a deep learning based approach to solve complex design exploration problems, specifically design of microfluidic channels for flow sculpting. A deep architecture is proposed to learn a sequence of actions that carries out the desired transformation over the input which has great implications on the innovation of manufacturing processes, material sciences, biomedical applications, decision planning, and many more. The creative integration of DL based tools can tackle the inverse fluid problem and achieve the required design accuracy while expediting the design process. Current efforts are primarily focusing on optimizing the tool-chain (e.g. rITN) as well as tailoring for specific application areas such as manufacturing and biology.

{\small
\bibliographystyle{ieee}

}

\end{document}